%% file: main.tex
\documentclass[letterpaper]{article}
\usepackage{aaai25}  % DO NOT CHANGE THIS
\usepackage{times}  % DO NOT CHANGE THIS
\usepackage{helvet}  % DO NOT CHANGE THIS
\usepackage{courier}  % DO NOT CHANGE THIS
\usepackage[hyphens]{url}  % DO NOT CHANGE THIS
\usepackage{graphicx} % DO NOT CHANGE THIS
\usepackage{natbib}  % DO NOT CHANGE THIS AND DO NOT ADD ANY OPTIONS TO IT
\usepackage{caption} % DO NOT CHANGE THIS AND DO NOT ADD ANY OPTIONS TO IT
\usepackage{algorithm}
\usepackage{algorithmic}
\usepackage{newfloat}
\usepackage{listings}
\usepackage{amsmath}
\usepackage{bm}
\usepackage{amssymb}
\usepackage{booktabs}
\usepackage{multirow}
\usepackage{mathrsfs}
\usepackage{url}
\usepackage{caption}
\usepackage{subcaption}
\usepackage{xspace}
\usepackage{color}
\usepackage[greek,english]{babel}

\newcommand{\ie}{\textit{i}.\textit{e}.}
\newcommand{\model}{STDN\xspace}

\DeclareCaptionStyle{ruled}{labelfont=normalfont,labelsep=colon,strut=off} % DO NOT CHANGE THIS
\floatstyle{ruled}
\newfloat{listing}{tb}{lst}{}
\floatname{listing}{Listing}
\newtheorem{Def}{Definition}

\title{Spatiotemporal-aware Trend-Seasonality\\Decomposition Network for Traffic Flow Forecasting}

\author{Lingxiao Cao, Bin Wang, Guiyuan Jiang, Yanwei Yu\thanks{Corresponding author: Yanwei Yu.}, Junyu Dong}
\affiliations {
    Faculty of Information Science and Engineering, Ocean University of China\\
    caolingxiao@stu.ouc.edu.cn, \{wangbin9545,jiangguiyuan,yuyanwei,dongjunyu\}@ouc.edu.cn
}
\begin{document}
\maketitle

\input{0abstract}
\input{1introduction}
\input{2relatedwork}
\input{3problem}
\input{4methodology}
\input{5experiment}
\input{6conclusion}

\clearpage
% \balance
\input{8Acknowledgment}

\bibliography{cite}

\clearpage
\end{document}

%% file: 0abstract.tex
\begin{abstract}

%Traffic flow prediction provides critical information for travel scheduling and public safety. However, the temporal variability and dynamic spatial changes of traffic data pose significant challenges for accurate forecasting. To capture the complex spatio-temporal heterogeneity, this paper proposes \textbf{\underline{S}}patiotemporal-aware \textbf{\underline{T}}rend-Seasonality \textbf{\underline{D}}ecomposition \textbf{\underline{N}}etwork (STDN) for traffic flow forecasting. Specifically, STDN incorporates graph structure learning and spatio-temporal embeddings learning to model and process sequence data and spatio-temporal data, respectively, and further explores the high-dimensional relationships between them. Subsequently, the data is divided into trend-cyclical and seasonal components in a simple but efficient manner for further processing. Finally, the data is encoded by an encoder for the two different types of data and decoded by a decoder. Particularly, since the existing real datasets are based on highways and do not adequately reflect the spatio-temporal changes within city, we release a new real-world dataset called JiNan to facilitate better development in this field. Extensive experiments conducted in real-world traffic prediction tasks demonstrate that \model achieves overall optimal performance, with great computational efficiency and data utilization. The source code of our method is available\footnote{https://github.com/xxxx}.

Traffic prediction is critical for optimizing travel scheduling and enhancing public safety, yet the complex spatial and temporal dynamics within traffic data present significant challenges for accurate forecasting. In this paper, we introduce a novel model, the  \textbf{\underline{S}}patiotemporal-aware \textbf{\underline{T}}rend-Seasonality \textbf{\underline{D}}ecomposition \textbf{\underline{N}}etwork (STDN). This model begins by constructing a dynamic graph structure to represent traffic flow and incorporates novel spatio-temporal embeddings to jointly capture global traffic dynamics. The representations learned are further refined by a specially designed trend-seasonality decomposition module, which disentangles the trend-cyclical component and seasonal component for each traffic node at different times within the graph. These components are subsequently processed through an encoder-decoder network to generate the final predictions. Extensive experiments conducted on real-world traffic datasets demonstrate that \model achieves superior performance with remarkable computation cost. Furthermore, we have released a new traffic dataset named JiNan, which features unique inner-city dynamics, thereby enriching the scenario comprehensiveness in traffic prediction evaluation.
%All source code and data are available. 
%\begin{links}
%\link{Code}{https://github.com/roarer008/STDN}
%\end{links}

%since the existing real datasets are based on highways and do not adequately reflect the spatio-temporal changes within a city,

\end{abstract}

%% file: 1introduction.tex
\section{Introduction}
\label{sec.intro}

With technological advancements, a diverse array of sensors has been increasingly integrated into monitoring systems to bolster modern intelligent transportation systems (ITS) \cite{cirstea2021enhancenet,ji2022stden, dai2023dynamic}. Transportation authorities deploy a variety of sensors, such as electronic police cameras and bayonet detectors, across road networks to continuously collect essential traffic data, including flow and speed. Utilizing historical traffic flow data and road network topology, traffic forecasting aims to predict future flow variations, thereby improving daily travel and traffic management \cite{dai2021temporal,li2023dynamic}.

In the realm of traffic forecasting, considerable efforts are devoted to modeling traffic dynamics. Methods based on deep learning, especially the Spatio-Temporal Graph Neural Networks (STGNNs), have proven more effective than statistical time series analysis and shallow machine learning techniques in addressing traffic forecasting challenges. To address temporal dynamics, sequential models like RNN-based variants \cite{graves2012LSTM,deng2022multi,deng2024disentangling} and non-sequential Transformers \cite{vaswani2017Transformer} have been profoundly studied. For spatial dynamics, recent progress has been made with the Graph Neural Networks (GNNs) \cite{yin2021survey}, which represent sensors as nodes within a graph, leveraging graph structure to capture traffic patterns. Despite these substantial advancements, our study suggests that current methods still exhibit considerable potential for improvement in two critical aspects.

% In the realm of traffic forecasting, considerable efforts have been devoted to modeling traffic dynamics. Deep learning have proven more effective than statistical time series analysis and shallow machine learning techniques in addressing traffic forecasting challenges. To address temporal dynamics, sequential models like RNN-based variants \cite{graves2012LSTM} and non-sequential Transformers \cite{vaswani2017Transformer} have been profoundly studied. For spatial dynamics, recent progress has been made with the Spatial Temporal Graph Neural Networks (STGNNs) \cite{yin2021survey}, which represent sensors as nodes within a graph, leveraging graph structure to capture traffic patterns. Despite these substantial advancements, our study suggests that current methods still exhibit considerable potential for improvement in two critical aspects.

%Firstly, data utilization efficiency is not enough, especially regarding the efficiency of using the unique temporal and spatial information in traffic series data. For the temporal aspects, traffic information is related to the specific week and time it occurs, as previous studies have also demonstrated \cite{guo2019ASTGCN}. However, prior works have merely coarsely inputted these two types of information into the models without fully integrating the temporal periodicity with the traffic information. For the spatial aspects, different regions have distinct spatial characteristics based on their locations. However, prior methods based on the graph constructed by distance between two points neglect deeper correlations among nodes in 

Firstly, the prominent spatio-temporal characteristics in traffic flow can be more effectively modeled with the appropriate inductive bias. Although previous studies have demonstrated \cite{zhang2017deep,yu2019citywide,guo2019ASTGCN}, traffic patterns are strongly influenced by the specific temporal periodicities, most efforts have roughly incorporated these temporal features into the models without explicitly modeling the synergy between the long and short periodicities \cite{chen2018price,deng2024parsimony}. Regarding spatial aspects, while different locations exhibit unique spatial characteristics, most GNN methods construct static graphs based solely on the distances between two nodes, fail to consider the global interactions among all nodes in the graph \cite{yin2021survey}.

Secondly, effective trend-seasonality decomposition of traffic flow can greatly enhance the representation learning of traffic nodes \cite{wu2021Autoformer, fang2023STWave}. Utilizing this methodology improves the prediction of traffic flow by distinguishing systematic patterns and noise components. However, the application of trend-seasonality decomposition predominantly to individual nodes in a traffic network overlooks the interactions among global nodes, thereby diminishing the quality of node representations learned by GNNs.
%Traffic flow in real word is complex and variable, with each observation containing rich high-order information related to its temporal and spatial context. The failure to model high-order spatio-temporal correlations hinders the node's representation learning.

%To address these two limitations, we propose a novel method named \textbf{\underline{S}}patio-\textbf{\underline{T}}emporal \textbf{\underline{A}}uto-\textbf{\underline{C}}orrelation Network (STAC) for traffic flow prediction, 

To bridge these research gaps, we introduce a \textbf{\underline{S}}patiotemporal-aware \textbf{\underline{T}}rend-Seasonality \textbf{\underline{D}}ecomposition \textbf{\underline{N}}etwork (STDN), which enhances global node representations through a novel trend-seasonality decomposition incorporating spatio-temporal embeddings. It features three key modules: \textit{(1) Module of Spatio-Temporal Embedding Learning} models the temporal periodicity by learning the temporal embedding including specific weeks and minutes, and acquires an initial spatial location embedding from the eigenvalues and eigenvectors of the graph Laplacian matrix. \textit{(2) Module of Dynamic Relationship Graph Learning} explores the global dynamic interaction among traffic nodes, enhanced by the spatio-temporal embedding, thereby capturing the high-order relationships between each traffic node. \textit{(3) Module of Trend-Seasonality
Decomposition} aims to refine the node representations by disentangling the traffic flow into the trend-cyclical and seasonal components, which are further processed through an encoder-decoder network. The contributions of our study are summarised as follows: %Extensive experiments on three traffic datasets demonstrate that our proposed STAC is capable of achieving superior performance compared with state-of-the-art methods in different settings.
%explores the higher-order spatio-temporal relationships across regions and times, we introduce a dynamic graph convolutional network, which incorporates graph structure learning to model neighborhood interactions. Then, a trend-seasonality decomposition module is proposed to 
% combines the vectors of neighborhood interaction with the spatio-temporal embedding

\begin{itemize}

    \item We propose the \model model, a novel dynamic GCN-based framework for traffic flow prediction. To our knowledge, this is the first approach to learn disentangled representations of traffic flow in view of the spatio-temporal embeddings.
    % \item We propose the \model model, a novel dynamic GCN-based framework for traffic flow prediction. To our knowledge, this is the first time of constructing dynamic graph that incorporates both spatio-temporal embeddings learning and trend-seasonality decomposition. 
    % \item We develop an innovative spatio-temporal embedding learning approach and a novel trend-seasonality decomposition mechanism. Each component of decomposition is designed to be aware of the spatio-temporal embeddings, enriching the model's ability to capture high-order interactions across time and space.
    \item We develop a novel trend-seasonality decomposition mechanism. Each component of decomposition is designed to be aware of the spatio-temporal embeddings, enriching the model's capability to capture high-order node interactions across time and space.
    \item We conduct comprehensive multi-step traffic flow prediction experiments on three real-world datasets. The experimental results demonstrate that our method consistently surpasses various competing baselines. Additionally, the effectiveness of each module is verified through the ablation study.
    \item We release a new urban dataset named JiNan, which, unlike popular highway traffic datasets (e.g., PeMS) focuses more on the spatio-temporal dynamics of inner-city traffic. We believe this dataset will further enrich the scenario comprehensiveness in traffic flow prediction evaluations.
    %We also publish a new urban dataset called JiNan, which includes real-time traffic flow information for various vehicles at different intersections. To advance developments in the transportation field.
    
\end{itemize}

%% file: 2relatedwork.tex
\section{Related Works}
We discuss related works from three categories of methods highly relevant to our study. % such as LSTM \cite{graves2012LSTM} and Transformer \cite{vaswani2017Transformer}. However, these approaches overlook the complex spatial correlations between sensors in the traffic network \cite{li2018DCRNN}.

\textbf{GCN-based Models}. Methods such as STGCN \cite{yu2017STGCN} and DCRNN \cite{li2018DCRNN} combine graph convolution network (\textbf{GCN}) with sequential data to model traffic flows. To address the limitations of predefined graphs based on the traffic network, GWNet \cite{wu2019GWNet} and AGCRN \cite{bai2020AGCRN} introduce adaptive graphs within GCNs, enhancing the capture of global and accurate spatial dependencies in traffic data. STGODE \cite{fang2021STGODE} utilizes tensor-based neural ODEs to mitigate the oversmoothing problem in deep GCNs. Subsequently, dynamic graph convolution \cite{han2021DMSTGCN, lan2022dstagnn, zhao2023DyHSL} are utilized to incorporate more intrinsic dynamic information within the spatial structure. However, they lose the guidance of prior knowledge, which may lead to underfitting and overfitting.

\textbf{Attention-based Models}. Attention-based models such as GMAN \cite{zheng2020GMANaaai} and ASTGCN \cite{guo2019ASTGCN} directly capture global temporal correlations between two time slices through attention mechanisms. Subsequent models like SSTBAN, STWave, and PDformer \cite{guo2023SSTBAN,fang2023STWave,jiang2023pdformer} extend this approach by integrating intrinsic spatial correlations with temporal dynamics through similar mechanisms. Although these models have achieved success in several aspects, they have not yet to fully leverage the potential of spatio-temporal embeddings within traffic sequence data.

\textbf{Decomposition-based Methods}. Decomposing time sequence data into the trend-cyclical and seasonal components has proven effective for prediction \cite{wu2021Autoformer,wang2024towards}. D\textsuperscript{2}STGNN \cite{D2STGNN} distinguishes two different types of hidden temporal signals: diffusion signals and intrinsic signals, which intriguingly parallel the trend-cyclical and seasonal parts. Other models treat seasonal part as delay signals \cite{jiang2023pdformer, long2024STDDE}. STWave \cite{fang2023STWave} adopts a unique approach by starting with the data itself, decomposing spatio-temporal data into trend and event components using wavelet transformations. However, these models overlook the unique characteristics across time and space.

% However, these models tend to overlook the intrinsic high-order interactions among nodes. In contrast, our model comprehensively addresses the complex correlations between spatio-temporal embeddings and traffic sequence data.

% However, these models overlook the intrinsic high-dimensional information across time and space. Compared to previous models, our model not only thoroughly considers the high-dimensional correlations between spatio-temporal embeddings and traffic sequence data, but also achieves higher accuracy with lower complexity.

%% file: 3problem.tex
\section{Problem Definition}
In this section, we define the key components and objectives of our study, focusing on the structure of traffic networks and the goals of traffic forecasting.

\begin{Def}[Traffic Network]
    Given the real-world traffic scenarios, we define the traffic network as a directed graph $\mathcal{G}=\{\mathcal{V},\mathcal{E},\mathbf{A}\}$, where $\mathcal{V}$ denotes a set of $N$ nodes, each corresponding to a different sensor within the road network. $\mathcal{E}$ represents a set of edges that denote the connectivity among the nodes. $\mathbf{A}$$\in$ $\mathbb{R}^{N\times N}$ is the adjacency matrix that models the connectivity between nodes.
\end{Def}

\begin{Def}[Traffic Forecasting]
    Given historical traffic time series and the road topology, the objective of traffic flow forecasting is to predict future values of traffic time series. Specially, we represent the historical time series as a signal tensor $\bm{\mathcal{X}} =$ $\left[\mathbf{X}_1,\mathbf{X}_2,\dots,\mathbf{X}_{T}\right]$ $\in$ $\mathbb{R}^{T\times N\times C}$, where $T$ is the length of historical traffic time series and $C$ is the number of dimensions of node attributes. We aim to construct a function $f(\cdot)$ that maps the historical time series over $T$ time steps to predict the subsequent $T^\prime$ time steps:
   \begin{equation}
        \left[\mathbf{X}_{1},\mathbf{X}_{2},\dots,\mathbf{X}_{T};\mathcal{G}\right] \stackrel{f(\cdot)}{\longrightarrow} \left[\mathbf{\hat{X}}_{T+1},\mathbf{\hat{X}}_{T+2},\dots,\mathbf{\hat{X}}_{T+T^\prime}\right].
    \end{equation}
\end{Def}

%% file: 4methodology.tex
\section{Methodology}

\begin{figure*}
    %\vspace{-4mm}
    \begin{center}
    \includegraphics[width=0.99\textwidth]{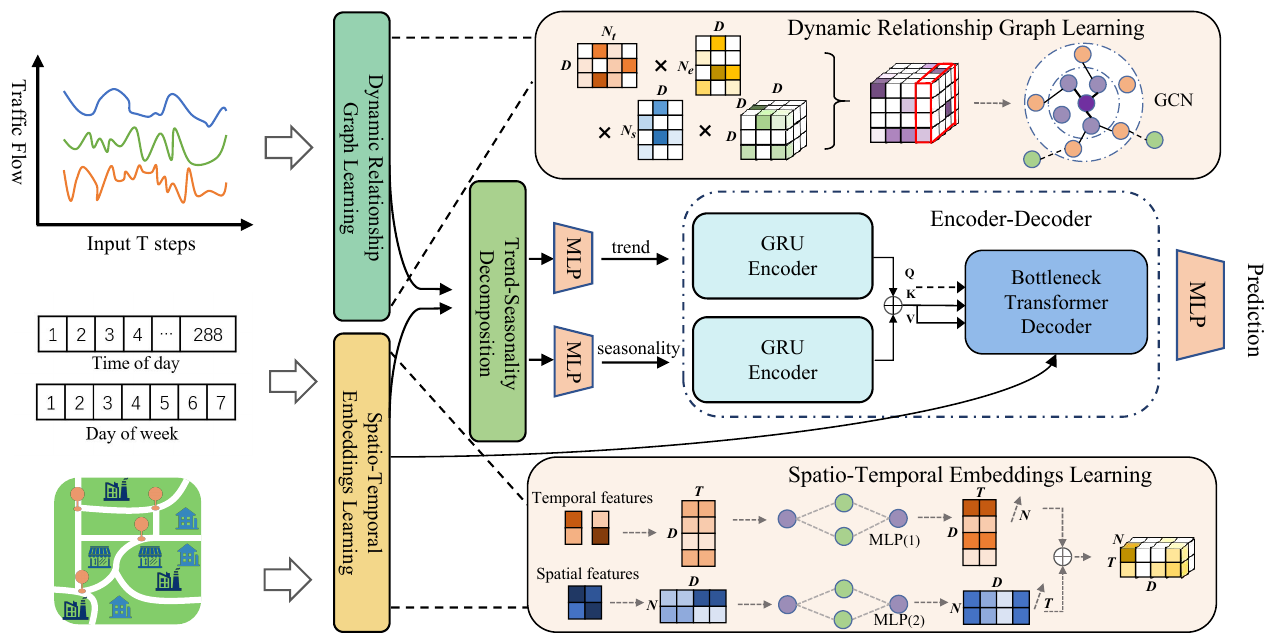}
    \caption{The overview of the proposed framework. MLP: multi-layer perceptron, GCN: graph convolution network.}
    \vspace{-6mm}
    \label{fig:overview}
    \end{center}
\end{figure*}

\model consists of three principal modules, which are introduced as follows. %Our \model first extends road information to spatio-temporal graphs that reference and deeply associate global nodes through dynamic relationship graph learning. Then, to better learn the inherent spatio-temporal information of the sequence data, we introduce a spatio-temporal embedding learning module. Finally, through a simple yet effective method, we combine the spatio-temporal embedding with the sequence data to generate trend-seasonality data, which are further processed through an encoder-decoder network to obtain the expected output.

\subsection{Module 1: Dynamic Relationship Graph Learning}
 %Previous methods mainly refer to graph constructed based on distance, without considering the high-order connectivity between each node. To model higher-order spatio-temporal relationships among all traffic nodes, we construct dynamic relationship graph considering different time steps and nodes. we first utilize a initial normalization and a dynamic relationship graph learning module, which adopts a GCN to explore neighborhood interactions.

 To address the limitations of simple distance-based connectivity, which overlooks the high-order relationships between each node, we construct a dynamic relationship graph that considers different time steps and nodes. This approach allows us to model the complex higher-order spatio-temporal relationships among all traffic nodes effectively.

Inspired by \citet{han2021DMSTGCN}, we design three learnable matrices and a learnable core tensor to streamline the constructing of the dynamic graph. These components include:  a time slot embedding $\mathbf{E}^{t} \in \mathbb{R}^{N_{t}\times D}$, a starting node embedding $\mathbf{E}^{s} \in \mathbb{R}^{N_{s}\times D}$, an ending node embedding $\mathbf{E}^{e} \in \mathbb{R}^{N_{e}\times D}$ and the core tensor $\bm{\mathcal{K}} \in \mathbb{R}^{D\times D\times D}$. Here, $N_{t}, N_{s}, N_{e}$ and the $D$ denote the number of time slots, starting nodes, ending nodes and the dimension of the embeddings, respectively. The specific calculations are as follows:
\begin{equation}
    \begin{aligned}
        \mathbf{A}_{t,i,j}^{\prime}&=\sum_{o=1}^D\sum_{q=1}^D\sum_{r=1}^D\bm{\mathcal{K}}_{o,q,r}\mathbf{E}_{t,o}^t\mathbf{E}_{i,q}^e{\mathbf{E}_{j,r}^s},\\
        \mathbf{A}_{t,i,j}^{\prime\prime}&=\max(0,\mathbf{A}_{t,i,j}^{\prime}),\\
        \mathbf{A}_{t,i,j}&=\frac{e^{\mathbf{A}_{t,i,j}^{\prime\prime}}}{\sum_{n=1}^{N_s}e^{\mathbf{A}_{t,i,n}^{\prime\prime}}}.
    \end{aligned}
\end{equation}

Through this methodology, we derive a tensor $\bm{\mathcal{A}} \in \mathbb{R}^{N_{t}\times N_{s}\times N_{e}}$ that encapsulates the high-order connectivity among global nodes across various time steps. Notably, since the starting nodes can also serve as ending nodes, we designate  $N = N_{s} = N_{e} $ for simplification.

Subsequently, to establish and analyze the relationships between different nodes, we employ a dynamic graph convolution. This process involves the matrix multiplication of the constructed adjacency matrix, the hidden states of the nodes, and the learnable parameters across different graphs at various time. The hidden states of the nodes are updated by aggregating the hidden states of their neighbors through weighted links at each time step:
\begin{equation}
    \bm{\mathcal{H}}_L=\sum\limits_{l=0}^L\left(\mathbf{A}_{(t)}\right)^l\mathbf{H}_{l}^t\mathbf{W}_l,
\end{equation}
where $\mathbf{A}_{(t)}$ denotes the adjacency matrix extracted from the tensor $\bm{\mathcal{A}}$ at time slot $t$. The term $(\mathbf{A}_{(t)})^l$ refers to the matrix multiplication operation. $\mathbf{H}_l^t$ represents the output hidden states at the $l$-th layer, which serves as input for the dynamic graph convolution in the $l$+1-th layer, $\mathbf{W}_l$ are the parameters specific to the $l$-th layer, and $\mathbf{H}_0^1$ represents the normalized traffic sequence data at first time step. Consequently, through these operations, we generate the final output $\bm{\mathcal{H}}_L \in \mathbb{R}^{T\times N\times D}$ which constitutes the processed sequence data.

% After convolution, we obtain the processed series data $\bm{\mathcal{\bar{X}}}$.
%After the convolution, we can get the series data from $\bm{\mathcal{X}}$ to $\bm{\mathcal{Y}}$ $\in$ $\mathbb{R}^{T\times N\times C}$.

\subsection{Module 2: Spatio-Temporal Embeddings Learning}
To more effectively capture the temporal correlations and periodicities in traffic flow, we design a temporal context embedding learning module. Given the time of day, represented as $\mathbf{z}^{d}_{h}$ $\in$ $\mathbb{R}^{T}$ and the day of the week, represented as $\mathbf{z}^{w}_{h}$ $\in$ $\mathbb{R}^{T}$, we extract temporal features and encode them by one-hot. By concatenating these encoded features (denoted by [,]), we generate an initial temporal embedding as below:
\begin{equation}
    \mathbf{Z}^{t}_{h}=\sigma(\mathbf{W}[onehot(\mathbf{z}^{d}_{h}), onehot(\mathbf{z}^{w}_{h})]),
    \label{eq:SEmbedding_1}
\end{equation}
where $\mathbf{Z}^{t}_{h}$ $\in$ $\mathbb{R}^{T\times D}$, $\sigma$ denotes the ReLU activation function. $\mathbf{W}$ comprises the trainable parameters. $\textit{D}$ specifies the dimensionality of the temporal embedding.

To enhance the ability of the temporal embedding to learn multi-resolution temporal features and to facilitate its integration with traffic time series, we refine the initial temporal embedding using a bias-free MLP:
\begin{equation}
    \mathbf{M}^{t}_{h}=\sigma_{2}(\mathbf{W}_{2}\sigma_{1}(\mathbf{W}_{1}\mathbf{Z}^{t})),
    \label{eq:SEmbedding_2}
\end{equation}
where $\sigma_{1}$ denotes the ReLU activation function, $\sigma_{2}$ is the sigmoid activation function, and $\mathbf{W}_{1}$, $\mathbf{W}_{2}$ are the trainable parameters. The resulting temporal embedding $\mathbf{M}^{t}_{h}$ $\in$ $\mathbb{R}^{T\times D}$ is utilized in subsequent stages of the model.

To effectively model the structure of the road network, we utilize the normalized Laplacian matrix, defined as:
\begin{equation}
    \Delta = \mathbf{I} - \mathbf{D}^{-1/2} \mathbf{A} \mathbf{D}^{-1/2},
\end{equation}
where $\mathbf{A}$ $\in$ $\mathbb{R}^{N\times N}$ is the adjacency matrix, $\mathbf{D}$ is the degree matrix, and $\mathbf{I}$ is the identity matrix. The Laplacian matrix's eigenvalues and eigenvectors encapsulate the spatial graph information in Euclidean space. By decomposing the Laplacian, we obtain the eigenvalue matrix and the eigenvector matrix:
\begin{equation}
    \Delta = \mathbf{U}\mathbf{\Lambda}\mathbf{U}^{\top},
\end{equation}
where $\mathbf{U}$ is the matrix of eigenvector  and $\mathbf{\Lambda}$ is the matrix of eigenvalue. To construct an initial spatial embedding, we select the $\textit{k}_{r}$ smallest nontrivial eigenvectors, resulting in $\mathbf{Z}^{s}$ $\in$ $\mathbb{R}^{N\times \textit{k}_{r}}$.
To capture global information as comprehensively as possible, we set $\textit{k}_{r}$ to 32.

To enhance the spatial embedding to represent spatial features and facilitate its integration with subsequent traffic time series, we process the initial spatial embedding using a bias-free MLP:
\begin{equation}
    \mathbf{M}^{s} = \mathbf{W}_{2}\sigma(\mathbf{W}_{1}\mathbf{Z}^{s}),
\end{equation}
where $\sigma$ is the ReLU activation function, $\textit{D}$ is the spatial embedding dimension, and $\mathbf{W}_1$, $\mathbf{W}_2$ are the trainable parameters. $\mathbf{M}^{s}$ $\in$ $\mathbb{R}^{N\times D}$ represents the spatial embedding that will be utilized in sebsequent stages of the model.

%It is noted that previous works merely coarsely input the initial temporal and spatial embedding vectors into the model \cite{jiang2023pdformer, li2024flashst} without effectively integrating the temporal and spatial embeddings with the sequence data. 
To effectively integrate the spatio-temporal embeddings with the traffic sequence data, we further propose a refined spatial-temporal embedding learning module. Specifically, this module involves broadcasting the spatial embedding $\mathbf{M}^{s}$ across the dimension $T$ to produce $\bm{\mathcal{M}}^{s} \in \mathbb{R}^{T\times N\times D}$, and similarly broadcasting the temporal embedding  $\mathbf{M}^{t}_{h}$ across the dimension $N$  to yield $\bm{\mathcal{M}}^{t}_{h} \in \mathbb{R}^{T\times N\times D}$. These embeddings are then combined as follows:
\begin{equation}
    \bm{\mathcal{M}} = \sigma_{1}(\bm{\mathcal{M}}^{t}_{h}) + \sigma_{2}(\bm{\mathcal{M}}^{s}),
    \label{eq:embeddings}
\end{equation}
where $\bm{\mathcal{M}}$ $\in$ $\mathbb{R}^{T\times N\times D}$, $\sigma_{1}$ represents the mathematical sine function, $\sigma_{2}$ denotes the ReLU activation function. 

In this manner, we obtain the refined spatio-temporal information embeddings for each node at different times. The use of the sine function is particularly beneficial as it not only normalizes the data to a range between -1 and 1 but also enables the model to effectively capture non-linear changes. Specifically, when we generate the initial temporal embedding $\mathbf{M}^{t}_{h}$, we employ the sigmoid activation function, as shown in equation \ref{eq:SEmbedding_2}. Consequently, after the application of the sine function, the data is normalized to a range between 0 and 1, which is crucial for effectively fusing the embedding with the traffic sequence data in subsequent processing steps.

\subsection{Module 3: Trend-Seasonality Decomposition}
Given the strong correlation between the trend-seasonality of a traffic node and its spatio-temporal context, the Trend-Seasonality Decomposition module is designed based on spatio-temporal embeddings learning. This module effectively disentangles the traffic flow of each node into distinct trend and seasonal components, adjusting the representations to better suit the forecasting task.
% allowing for a precise modeling of traffic dynamics.
%Unlike previous models that directly use a single sequential method \cite{wu2021Autoformer, fang2023STWave} or a mathematical method to model traffic sequence data , we decompose traffic series data into trend-cyclical and seasonal components by the spatio-temporal embeddings. Clearly, the temporal variations of trend-cyclical data and seasonal data are markedly different. Trend-cyclical data exhibits stable and continuous temporal changes, whereas seasonal data shows fluctuating and delay changes. We believe that trend-cyclical data is highly correlated with the temporal and spatial context it resides in. Intuitively, a node exhibits different trend changes at different times, hence seasonal data can be derived from its own spatio-temporal variations.
% Due to the fluctuation of spatial information in traffic sequence data, the trend changes between nodes are also different. 
% Through the spatio-temporal embeddings learning, we obtain trend changes for different regions at different time. 

When processing the traffic sequence data with trend-seasonality decomposition, since the dimensionality of the traffic hidden states and the spatio-temporal embeddings has been aligned, we initially derive the trend component by through multiplication-wise interaction between the hidden states of the nodes $\bm{\mathcal{H}}_L$ and the spatio-temporal embeddings $\bm{\mathcal{M}}$. The residual part constitutes the seasonal component. The calculations are as follows:
\begin{equation}
    \begin{aligned}
        \bm{\mathcal{X}}_{t}& = \bm{\mathcal{H}}_L \odot \bm{\mathcal{M}}, \\ 
        \bm{\mathcal{X}}_{s}& = \bm{\mathcal{H}}_L - \bm{\mathcal{X}}_{t},
    \end{aligned}
\end{equation}
where $\bm{\mathcal{X}}_{s}, \bm{\mathcal{X}}_{t} \in \mathbb{R}^{T\times N\times D}$ denote the seasonal and the extracted trend-cyclical part respectively. $\odot$ is the Hadamard product. The trend component of traffic flow is generally smooth and strongly influenced by the time and location of the traffic node. Our approach is based on the mild assumption that nearby times and locations yield similar trends, while distinct times and locations exhibit unique traffic patterns.

\subsection{Encoder-Decoder Architecture}
In this module, we utilize an encoder-decoder network to extract deeper spatio-temporal features. For the encoding process, we employ Gated Recurrent Unit (GRU) due to its effectiveness in capturing temporal dependencies. For the decoding process, we opt for the Transformer architecture, primarily because of its unique multi-head attention mechanism, which generates predictions with superior performance. We denote use the GRU($\cdot$) to denote the Gated Recurrent Unit as below:
\begin{equation}
    \begin{aligned} 
        &\bm{\mathcal{Y}}_{t} = GRU(\bm{\mathcal{X}}_{t}), \\
        &\bm{\mathcal{Y}}_{s} = GRU(\bm{\mathcal{X}}_{s}), 
    \end{aligned}
\end{equation}
where $\bm{\mathcal{Y}}_{s}, \bm{\mathcal{Y}}_{t} \in \mathbb{R}^{T\times N\times D}$ represent the outputs of the GRU encoder for the seasonal and trend components, respectively. These outputs are then combined through element-wise summation to produce the final output $\bm{\mathcal{Y}} \in \mathbb{R}^{T\times N\times D}$.

The Transformer architecture utilizes multi-head attention mechanisms, which first projects the queries, keys and values into $h$ different $d$-dimensional subspaces, and then execute the attention function in parallel:
\begin{equation}
    \begin{aligned}
        \mathrm{MHSA(Q,K,V)}& =\oplus(\mathrm{head}_1,\ldots,\mathrm{head}_h)W^O, \\
        \mathrm{head}_{j}& =\mathrm{softmax}(\frac{(\mathbf{Q}W_{j}^{Q})(\mathbf{K}W_{j}^{K})^{\top}}{\sqrt{d}})(\mathbf{V}W_{j}^{V}), 
    \end{aligned}
\end{equation}
where $W$ is the training parameter.

Inspired by \citet{guo2023SSTBAN}, we incorporate a component known as the Bottleneck Transformer Block (BT block) into our architecture, strategically designed to reduce both temporal and spatial complexity.

For the predicted time of day and day of the week, we obtain the predicted time embedding $\mathbf{M}^{t}_{p}$ $\in$ $\mathbb{R}^{T^\prime \times D}$ according to equation \ref{eq:SEmbedding_1} and equation \ref{eq:SEmbedding_2}. To align with the default settings, we maintain $T = T^\prime$. Additionally, we extend the temporal embedding $\mathbf{M}^{t}_{p}$ across the dimension $ N $, resulting in $\bm{\mathcal{M}}^{t}_{p} \in \mathbb{R}^{T^\prime\times N\times D}$. 

To effectively prompt the decoder to focus on capturing the high-order dynamics between time and space, we concatenate the predicted temporal embedding and spatial embedding with the output from $(l - 1)^{th}$ BT block as $\bm{\mathcal{Z}}^{(l-1)} \in \mathbb{R}^{T^\prime\times N\times D}$ to obtain $\bm{\mathcal{H}}^{(l)}=(\bm{\mathcal{Z}}^{(l-1)}||\bm{\mathcal{M}}^{t}_{p}||\bm{\mathcal{M}}^{s}) \in \mathbb{R}^{T^\prime \times N \times 3D}$. Besides, $\bm{\mathcal{H}}^{(l-1)}_{:,v} \in \mathbb{R}^{T^\prime\times 3D}$ represents the input pertaining to node $v$ across all time slices.

The specific BT block employs defined $T^\prime$ vectors $\mathbf{IT}^{(l)} \in \mathbb{R}^{T^\prime\times 3D}$ as trainable parameters within the model. The transformation within the BT block is described by the following equations:
\begin{equation}
    \begin{aligned}
        \mathbf{IT}^\prime &= \mathrm{MHSA}(\mathbf{IT}^{(l)},\bm{\mathcal{H}}^{(l-1)}_{:,v},\bm{\mathcal{H}}^{(l-1)}_{:,v}) \in \mathbb{R}^{T^\prime\times 3D},\\
       \bm{\mathcal{Z}}^{(l)}_{:,v} &= \mathrm{MHSA}(\bm{\mathcal{H}}^{(l-1)}_{:,v},\mathbf{IT}^\prime,\mathbf{IT}^\prime) \in \mathbb{R}^{T^\prime\times 3D}.
    \end{aligned}
\end{equation}
The initial input to the first BT block is $\bm{\mathcal{Y}}$ sourced from the encoder. The output from the final $l$ BT block is denoted as $\bm{\mathcal{Z}} \in \mathbb{R}^{T^\prime\times N\times D}$, which represents the projected future representation of traffic flow.

Ultimately, to derive the expected traffic predictions, we employ a fully-connected neural network that transforms the future representation of traffic flow into the predicted values $\bm{\mathcal{\hat{X}}} \in \mathbb{R}^{T^\prime\times N\times C}$. We utilize the $L1$ as the training loss function as below:
\begin{equation}
    \mathcal{L} = \sum_{t=T+1}^{T+T^\prime}\sum_{n=1}^{N}\lvert{\hat{x}_{t}^{n}} - y_{t}^{n}\rvert.
    \label{eq:loss}
\end{equation}

%% file: 5experiment.tex
\section{Experiments}
\input{table/datasets.tex}

\subsection{Datasets}

%We utilize three real-world datasets to evaluate the performance of \model. The JiNan dataset is derived from traffic flow statistics in a real city. Similar to the processing in \cite{song2020STSGCNaaai}, we set the time interval to 5 minutes. However, unlike \cite{song2020STSGCNaaai}, our constructed adjacency matrix is not based on distance but on connectivity. Specifically, if two detection points are on the same road, an edge is constructed between them. The remaining two datasets are based on California's highways, which differ geographically from JiNan dataset. The detailed information is shown in Table~\ref{tab.datasets}.

To evaluate the performance of \model, we utilize three real-world datasets, each offering unique traffic flow dynamics. The dataset JiNan, which is first released by us, derived from actual traffic flow statistics in a city, mirrors the setup in \cite{song2020STSGCNaaai}, where the time interval is set to 5 minutes. 
% When we construct the experimental dataset, we establish edges between neighbor detection points if they are connected on the road. The other two datasets PeMS providing a geographical contrast to the JiNan dataset. 
Detailed descriptions of these datasets are provided in Table~\ref{tab.datasets}.

\input{table/perform_compared}

\subsection{Baseline Methods}

To assess the performance of \model, we compare it against a diverse range of established baselines:

\begin{itemize}
    \item \textbf{HA} \cite{hamilton2020HA} uses the historical average 
    of input data for prediction.
    \item \textbf{ARIMA} \cite{box2015ARIMA} is a well-known statistical model widely employed for time series forecasting.
    \item \textbf{VAR} \cite{lutkepohl2005VAR}: is another traditional method for time series forecasting.
    \item \textbf{SVR} \cite{wu2004SVR} employs support vector regression for predictive modeling.
    \item \textbf{LSTM} \cite{graves2012LSTM} is a deep learning model that captures temporal dependencies but does not account for spatial correlations.
    \item \textbf{DCRNN} \cite{li2018DCRNN} integrates diffusion convolution into GRU layers to enhance spatial-temporal correlation capture.
    \item \textbf{STGCN} \cite{yu2017STGCN} combines graph and temporal convolutions to handle spatial-temporal data.
    \item \textbf{GWNet} \cite{wu2019GWNet} combines dilated convolution with diffusion graph convolution and introduces a self-adaptive adjacency matrix.
    \item \textbf{GMAN} \cite{zheng2020GMANaaai} employs multi-attention to capture both spatial and temporal dynamics.
    \item \textbf{ASTGCN} \cite{guo2019ASTGCN} applies attention mechanisms on both temporal and spatial convolutions to dynamically capture spatio-temporal correlations.
    \item \textbf{AGCRN} \cite{bai2020AGCRN} focuses on extracting node-specific features and uncovering hidden node interdependencies.
    % \item \textbf{ASTGNN} \cite{guo2021lASTGNN}: designs trend-aware self-attention module and a dynamic graph convolution module to capture spatialtemporal dynamics.
    \item \textbf{DMSTGCN} \cite{han2021DMSTGCN}  learns dynamic spatial dependencies and builds a multi-faceted fusion module for complex traffic data features.
    \item \textbf{STGODE} \cite{fang2021STGODE} adopts ordinary differential equations for traffic flow forecasting.
    \item \textbf{STGNCDE} \cite{choi2022STGNCDE} designs two neural controlled differential equations for prediction.
    \item \textbf{D\textsuperscript{2}}\textbf{STGNN} \cite{D2STGNN} models traffic flow by separating it into the diffusion component and the inherent component.
    \item \textbf{DSTAGNN} \cite{lan2022dstagnn} constructs a spatio-temporal graph and utilizes multi-head attention to represent dynamic spatial relevance.
    \item \textbf{SSTBAN} \cite{guo2023SSTBAN} implements a self-supervised learning approach with a masking method for prediction.
    \item \textbf{STWave} \cite{fang2023STWave} employs wavelets to decompose traffic data into stable trends and fluctuating events.
\end{itemize}

\subsection{Evaluation Metrics and Experimental Settings}
In our evaluation,  we employ the mean absolute error (MAE), root mean square error (RMSE) and mean absolute percentage error (MAPE) to quantify the performance of different methods. 

Our experiments are conducted on a server with NVIDIA RTX 4090 GPU cards, running CUDA version 12.2. All the models are implemented using PyTorch.
The datasets are split in a 6:2:2 ratio for training, validation, and testing, respectively. We use historical data from the past hour to predict the traffic flow for the next hour, corresponding to using the past 12 time steps to forecast the next 12 steps.

To prevent overfitting, an early-stopping strategy is employed with a patience setting of 10. We use Adam optimizer with an initial learning rate of 0.001. The standard batch size for all experiments is set to 64. If GPU memory constraints occur, the batch size is reduced to 32, and further to 16 if necessary, until the programs can run efficiently. The number of dimensions of node attribute on three datasets is $C$ = 1. Totally, there are 3 hyperparameters in our model, \ie, the numbers of bottleneck transformer block $L$, the number of attention heads $h$, and the dimensionality $d$ of each attention head, where the total number of features $D = h \times d$. The optimal settings for our model  on PeMS04 and JiNan datasets are $L = 2$, $h = 8$, $d = 16$ ($D = 128$). For the PeMS07 dataset, the best performance is achieved with $L = 2$, $h = 8$, $d = 12$ ($D = 96$). All source code and data are available at \url{https://github.com/roarer008/STDN}

%6:2:2 超参设置，比如d,k,l,adam,batch_size,GPU,learning rate
\subsection{Performance Comparison}
Table~\ref{tab:perform_compared} presents the results from graph-based baselines and grid-based baselines. The best results are highlighted in bold, and the second-best results are underlined. Based on these results, several key conclusions can be drawn:
\begin{itemize}
    \item \model achieves state-of-the-art performance, particularly evident in the PeMS04 and JiNan datasets. Traditional machine learning methods such as ARIMA typically perform poorly, as they are unable to capture the non-linear correlations present in the spatio-temporal traffic data.%With the exception of the MAPE metric, \model  surpasses other baselines on the PeMS07 dataset. %We posit that GRU is more beneficial for less-nodes datasets. 
    
    %\item Traditional machine learning methods such as ARIMA typically perform poorly, as they are unable to capture the non-linear correlations present in the spatio-temporal traffic data.
    
    \item  Among the GCN-based models, AGCRN demonstrates strong performance. Compared to other models, \model excels in capturing the structure of the road network by effectively integrating eigenvalues from the Laplacian matrix with traffic flow data. 
    %Besides that, attention-based models generally perform near optimally among all baselines. Notably, SSTBAN demonstrates strong performance.
    \item Attention-based models generally perform near optimally among all baselines. Notably, \model distinguishes itself by integrating spatio-temporal embeddings with traffic flow data and the decoder, significantly enhancing traffic forecasting accuracy.
    \item Compared to the baseline models, we incorporate multi-resolution temporal features, such as "time of day" and "day of week", for temporal embeddings, alongside a geospatial directed graph for spatial embeddings. Based on these spatiotemporal embeddings, traffic flow is disentangled into trend and seasonality parts. This novel disentangling method significantly enhances the traffic forecasting accuracy.
    
\end{itemize}
% (1) \model achieves state-of-the-art performances and the advantages are more evident in the PeMS04 and JiNan datasets. We argue that GRU is more beneficial for less-nodes datasets. (2) Among the GCN-based models, AGCRN performs well. Compared to other GCN-based models, \model effectively captures the road network's structure by efficiently integrating eigenvalues from the Laplacian matrix with traffic series data. (3) Among the attention-based models, GMAN and DSTAGNN are the best baselines. Compared to other attention-based models, \model excels by incorporating both spatial and temporal information in the encoder’s objective for prediction.

\subsection{Ablation Study}

\input{table/ablation}
To evaluate the effectiveness of different components in \model, we conducted the ablation study with several variants of the \model:
\begin{itemize}
    \item \textbf{w/o TE}: This variant removes the temporal embedding modeling, meaning the decoder operates solely with spatial embedding cues.
    \item \textbf{w/o SE}: This variant removes the spatial embedding modeling, meaning the decoder operates solely with temporal embedding cues.
    \item \textbf{w/o STE}: This variant eliminates spatio-temporal embedding, thus the traffic flow is not decomposed into trend-seasonality components. As a result, the decoder does not incorporate any spatio-temporal cues.
    \item \textbf{w/o DRG}: This variant eliminates the dynamic relationship graph learning.
    \item \textbf{w/o STD}: Instead of using spatiotemporal-aware decomposition to disentangle the traffic sequence data, this variant adopts the decomposition method utilized by Autoformer \cite{wu2021Autoformer}.
\end{itemize}
% (1) \textit{w/o TE}: this variant removes the temporal embedding modeling, meaning the decoder operates solely with spatial embedding cues. (2) \textit{w/o SE}: this variant removes the spatial embedding modeling, meaning the decoder operates solely with temporal embedding cues. (3) \textit{w/o STE}: this variant eliminates spatio-temporal embedding, thus the traffic flow is not decomposed into trend-seasonality components. As a result, the decoder does not incorporate any spatio-temporal cues. (4) \textit{w/o DRG}: this variant removes the dynamic relationship graph learning. (5) \textit{w/o STD}: instead of using spatiotemporal-aware decomposition to disentangle the traffic sequence data, this variant adopts the decomposition method utilized by Autoformer \cite{wu2021Autoformer}.

Table~\ref{tab.ablation} presents the comparison of \model and its variants on PeMS04 and JiNan datasets. From this comparison, we can draw several conclusions: (1) The origin \model consistently achieves the best performance relative to its variants, underscoring the effectiveness of its full configuration. (2) The results show that the variant ``\textit{w/o TE}'' generally outperforms ``\textit{w/o SE}'' across most tasks. This suggests that temporal information, particularly periodic information, plays a more critical role than spatial information. (3) The ``\textit{w/o DRG}'' underperforms \model, indicating the importance of the dynamic relationship graph learning module. (4) The performance of ``\textit{w/o STD}'' underlines the necessity of the trend-seasonality decomposition module aware of spatio-temporal embeddings.
% \input{table/ablation}
% Table~\ref{tab.ablation} shows the comparison of these variants on the PeMS04 and JiNan datasets. Based on the results, we can draw the following conclusions: (1) The results show the superiority of SSA over GCN in capturing dynamic and long-range spatial dependencies. (2) PDFormer leads to a large performance improvement over w/o Mask, highlighting the value of using the mask matrices to identify the significant node pairs. In addition, w/o SemSAH and w/o GeoSAH perform worse than PDFormer, indicating that both local and global spatial dependencies are significant for traffic prediction. (3) w/o Delay performs worse than PDFormer because this variant ignores the spatial propagation delay between nodes but considers the spatial message passing as immediate.
\vspace{-2mm}
\subsection{Parameter Sensitivity Study}
\input{figure/ablation}
Figure \ref{fig:abla} illustrates the results of hyper-parameter sensitivity analysis for our \model on PeMS04 and PeMS07 datasets. This study involved varying the number of decoder layers and the number of features in \model, exploring options within the ranges of [1, 2, 3, 4] for layers and [32, 64, 96, 128, 160] for features. From this analysis, we can draw several conclusions: (1) The performance of our model improves with an increasing in the number of decoder layers but stabilizes at 2 layers. (2) Optimal performance is achieved with 128 features on the PeMS04 dataset and 96 features on the PeMS07 dataset. This finding highlights that while increasing the number of features generally enhances the model's capability to represent complex traffic patterns, but excessive features may introduce noise and degrade model performance.

\subsection{Model Efficiency Study}
\begin{figure}
    %\vspace{-5mm}
    \begin{center}
    \includegraphics[width=0.46\textwidth]{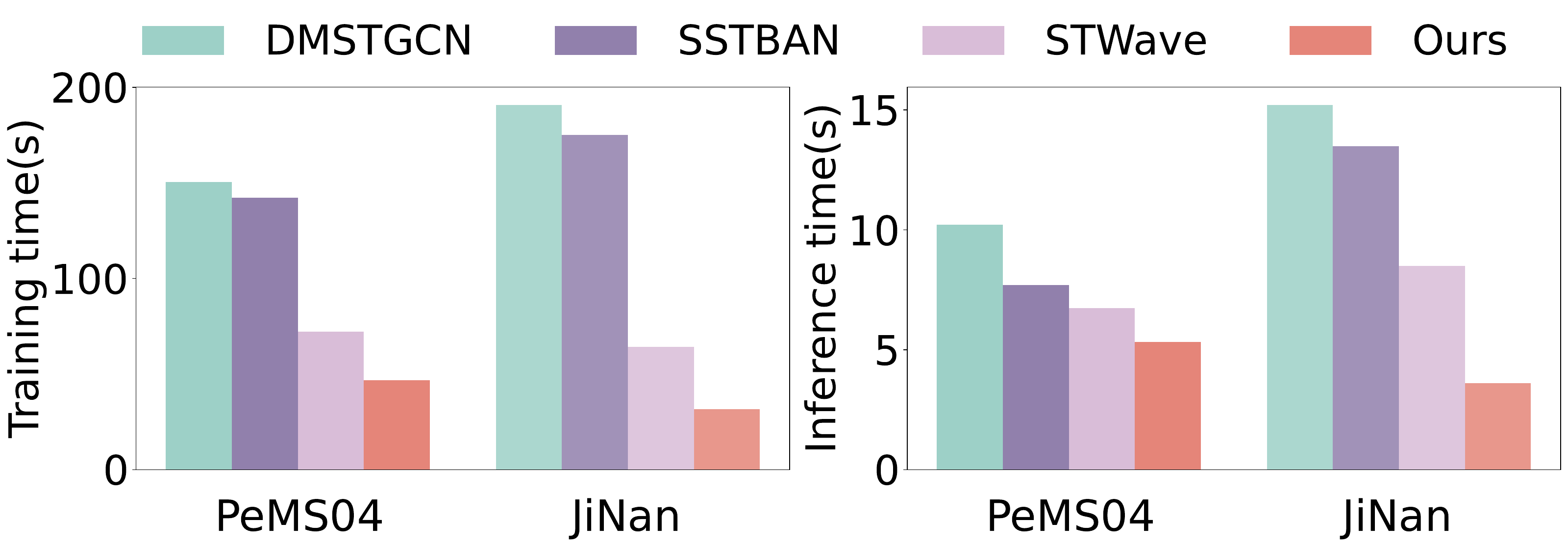}
    %\vspace{-4mm}
    \caption{The computational time cost on PeMS04 and JiNan datasets.}
    \vspace{-3mm}
    \label{fig:time}
    \end{center}
\end{figure}
\begin{figure}
    \centering
    \vspace{-1mm}
    \includegraphics[width=0.98\linewidth]{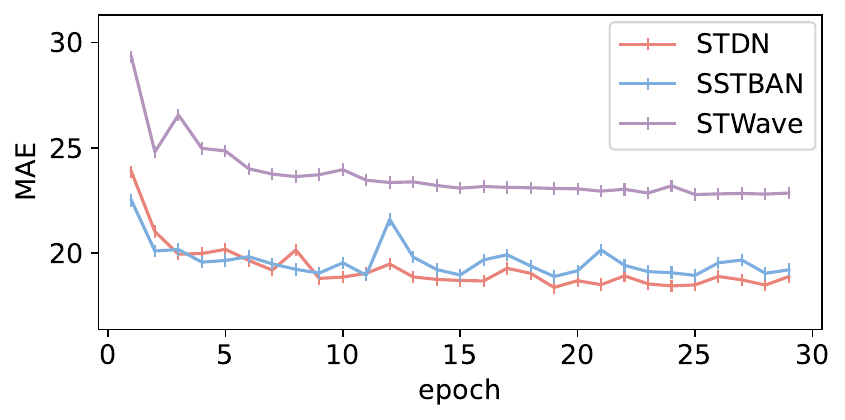}
    \vspace{-0.5mm}
    \caption{The MAE on the validation part of PeMS04 dataset during the training process.}
    \vspace{-4mm}
    \label{fig:epoch}
\end{figure}
To demonstrate the efficiency of our model, we benchmark \model against DMSTGCN, SSTBAN, and STWave, which have achieved suboptimal results at the PeMS04 and JiNan datasets. Figure~\ref{fig:time} displays the average training time per epoch and inference time for each model. Figure~\ref{fig:epoch} illustrates the MAE curves on the validation part of PeMS04 dataset during the training process. The following observations can be made: (1) \model not only trains faster but also infers quicker than the compared models. (2) \model demonstrates a faster convergence rate, achieving better performance in fewer epochs. In our experiments, while STWave reaches its best performance at epoch 100, its MAE is still higher than the lowest MAE achieved by our \model at just epoch 19. 

The computational complexity of our \model encoder-decoder module is $O(TD + LND)$, with the encoder and decoder contributing complexities of $O(TD)$ and $O(LND)$, respectively. Here, $L$ denotes the number of bottleneck transformer blocks. The complexity of the spatio-temporal embedding module is given by $O((T + N)D + N^3)$. Although calculating the eigenvectors and eigenvalues of the graph Laplacian is computationally intensive, marked by a complexity of  $O(N^3)$, this process can be efficiently handled through preprocessing prior to training. Therefore, \model maintains comparable time and memory complexity during training, ensuring efficiency without compromising performance.

%% file: table/datasets.tex
\begin{table}
    \begin{center}
    % \vspace{-2mm}
    % \fontsize{9}{12} \selectfont
    \setlength{\tabcolsep}{1mm}{}	
    \begin{tabular}{ccccc}
        \toprule
        \textbf{Datasets}&Nodes&Edges&Time Interval&Time range \\
        \midrule
        \text{PeMS04} & 307 & 340 & 5\text{min} & 01/2018-02/2018 \\
        \text{PeMS07} & 883 & 866 & 5\text{min} & 05/2017-08/2018 \\
        \text{JiNan} & 406 & 324 & 5\text{min} & 08/2016-09/2016 \\
        % \text{SD} & 716 & 17,319 & 15\text{min} & 01/2019-12/2019 \\
        % \text{GBA} & 2,352 & 61,246 & 15\text{min} & 01/2019-12/2019 
        \bottomrule
    \end{tabular}
    \vspace{-2mm}
    \caption{Statistics of the datasets.}
    \vspace{-8mm}
    \label{tab.datasets}
    \end{center}
\end{table}

%% file: table/perform_compared.tex
\begin{table*}[t]

%\vspace{-6mm}
\centering

% \vspace{-2mm}
%\footnotesize

\fontsize{9.5}{10} \selectfont
\setlength{\tabcolsep}{3.5mm}{}

\begin{tabular}{cccccccccc}
\toprule
\midrule
\multirow{2}{*}{Model} & \multicolumn{3}{c}{PeMS04} & \multicolumn{3}{c}{PeMS07} & \multicolumn{3}{c}{JiNan}\\
%\cline{2-10}
% & MAE & RMSE & MAPE & MAE & RMSE & MAPE & MAE & RMSE & MAPE\\
\cmidrule(lr){2-4} \cmidrule(lr){5-7} \cmidrule(lr){8-10} % 使用 cmidrule 替代 cline
 & \raisebox{-0.1ex}{MAE} & \raisebox{-0.1ex}{RMSE} & \raisebox{-0.1ex}{MAPE} 
 & \raisebox{-0.1ex}{MAE} & \raisebox{-0.1ex}{RMSE} & \raisebox{-0.1ex}{MAPE} 
 & \raisebox{-0.1ex}{MAE} & \raisebox{-0.1ex}{RMSE} & \raisebox{-0.1ex}{MAPE} \\
\midrule

HA &38.03  &59.24  &27.88\%  &45.12 &65.64 &24.51\%  &13.23 &22.85 &46.83\% \\
ARIMA &33.73  &48.80  &24.18\%   &38.17 &59.27 &19.46\% &14.89 &26.22 &48.53\% \\
VAR &24.54  &38.61  &17.24\%    &50.22 &75.63 &32.22\% &12.56 &20.60 &40.54\% \\
SVR &28.70  &44.56  &19.20\%   &32.49 &50.22 &14.26\%  &11.98 &20.82 &39.56\%\\
LSTM &26.77  &40.65  &18.23\%  &29.98 &45.94 &13.20\% &11.30 &19.40 &38.02\%\\
DCRNN &21.02  &33.44  &14.17\%   &25.22 &38.61 & 11.82\% &11.40 &20.54 &41.74\%\\
STGCN &21.16  &34.89  &13.83\%   &25.33 &39.34 &11.21\% &9.41 &16.08 &36.42\%\\
GWNet &24.89  &39.66  &17.29\%   &26.39 &41.50 &11.97\% &9.39 &16.13 &35.83\%\\
GMAN &19.14  &31.60  &13.19\%   &20.97 &34.02 &9.05\%  &10.15 &16.98 &38.32\%\\
ASTGCN(r) &22.93  &35.22  &16.56\%   &24.01 &37.87 &10.73\% &10.06 &17.23 &38.95\%\\
AGCRN &19.83  &32.26  &12.97\%  &22.37 &36.55 &9.12\%  &9.52 &16.15 &40.27\%\\
% ASTGNN &18.66  &31.13  &12.47\%  &0.7481  &0.7562  &0.7633  \\
DMSTGCN  &20.01 &32.18 &14.50\%  &23.73 &36.01 &12.21\% &8.78 &\underline{14.61} &34.52\%\\
STGODE &20.84  &32.82  &13.77\%   &22.59 &37.54 &10.14\% &9.25 &15.75 &37.68\%\\
STGNCDE &19.21  &\underline{31.09}  &12.76\%   &20.53 &\underline{33.84} &\underline{8.80\%} &9.14 &16.60 &36.89\%\\
D\textsuperscript{2}STGNN &19.55  &31.99  &12.82\%    &21.55 &34.83 &9.39\% &9.12 &15.97 &35.65\%\\
DSTAGNN &19.30 &31.46 &12.70\% &21.42 &34.51 &9.01\% &8.95 &14.99 &36.87\% \\
\midrule
% PDFormer & 18.32  &29.97  & 12.10\%  &{0.8216}  &{0.8032}  &{0.8156}  \\ 
SSTBAN &\underline{18.89}  &31.21  &\underline{12.62\%}    &20.45 &37.55 &9.31\% &\underline{8.41}	&14.62	&\underline{33.79\%}\\ 
STWave &21.20  &34.47  &14.32\%   &\underline{20.33} &34.03 &\textbf{8.58\%} &9.23 &16.10  &36.12\%\\
\midrule
\textbf{\model} &\textbf{18.40$^*$}	&\textbf{30.41$^*$}	&\textbf{12.21\%$^*$}    &\textbf{20.08$^*$}	&\textbf{33.73$^*$}	&9.29\% &\textbf{8.32$^*$}	&\textbf{14.52$^*$}	&\textbf{32.19\%$^*$}\\
% \textbf{\model(Transformer)} &19.46	&31.40	&14.75\%  &8.67	&15.03	&37.25\% &18.38	&31.31	&12.11\%  \\
\midrule
\bottomrule
\end{tabular}
\vspace{-1mm}
\caption{Performance comparison of all models on three real-world datasets. Marker $*$ indicates the
results are statistically significant (t-test with p-value $<$ 0.01).}
\label{tab:perform_compared}
\vspace{-6mm}
\end{table*}

%% file: table/ablation.tex
\begin{table}
    \begin{center}
    
    % \vspace{-2mm}
    % \fontsize{9}{12} \selectfont
    \setlength{\tabcolsep}{0.8mm}{}	
    \begin{tabular}{ccccccc}
        \toprule
        \multirow{2}{*}{Model} & \multicolumn{3}{c}{PeMS04} & \multicolumn{3}{c}{JiNan} \\
        %\cline{2-7}
        \cmidrule(lr){2-4} \cmidrule(lr){5-7} 
 & \raisebox{-0.1ex}{MAE} & \raisebox{-0.1ex}{RMSE} & \raisebox{-0.1ex}{MAPE} 
 & \raisebox{-0.1ex}{MAE} & \raisebox{-0.1ex}{RMSE} & \raisebox{-0.1ex}{MAPE} \\
        %& MAE & RMSE & MAPE & MAE & RMSE & MAPE \\
        \midrule
        \text{w/o TE} &19.10 &30.89 &12.87\% &8.47	&14.67	&34.18\% \\
        \text{w/o SE} &18.55 &30.46 &12.52\% &8.44	&14.68	&35.15\% \\
        \text{w/o STE} &18.93 &30.70 &12.97\% &8.55	&14.79	&35.38\% \\
        \midrule
        \text{w/o DRG} &18.61 &30.90 &12.23\% &8.37	&14.57	&33.91\% \\
        \text{w/o STD} &18.60 &31.08 &12.33\% &8.75	&15.14	&37.88\% \\
        \midrule
        \text{\model} &\textbf{18.40}	&\textbf{30.41} &\textbf{12.21\%}  &\textbf{8.32}	&\textbf{14.52}	&\textbf{32.19\%}\\
        \bottomrule
    \end{tabular}
    \vspace{-2mm}
    \caption{Ablation study on PeMS04 and JiNan datasets.}
    \label{tab.ablation}
    \vspace{-8mm}
    \end{center}
\end{table}

%% file: figure/ablation.tex
\begin{figure}
    \centering
    %\vspace{-3mm}
    \begin{subfigure}{0.23\textwidth}
        \includegraphics[width=\linewidth]{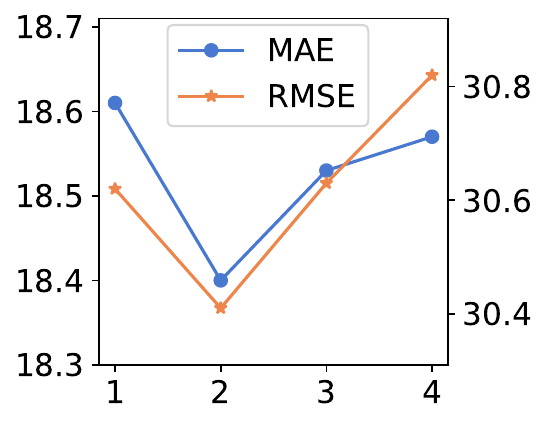}
        \vspace{-6mm}
        \caption{\#Layers on PeMS04}
        \label{fig:recall_dim}
    \end{subfigure}
    \begin{subfigure}{0.22\textwidth}
        \includegraphics[width=\linewidth]{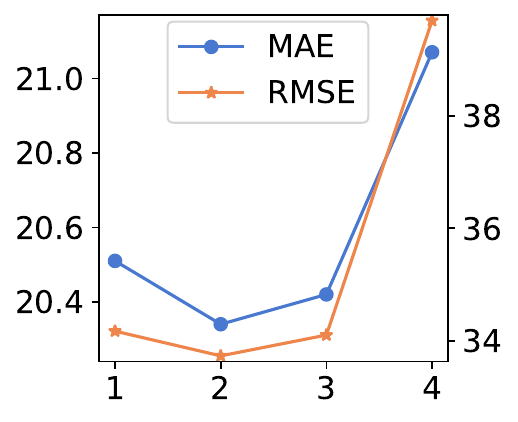}
        \vspace{-6mm}
        \caption{\#Layers on PeMS07}
        \label{fig:ndcg_dim}
    \end{subfigure}
    \begin{subfigure}{0.23\textwidth}
        \includegraphics[width=\linewidth]{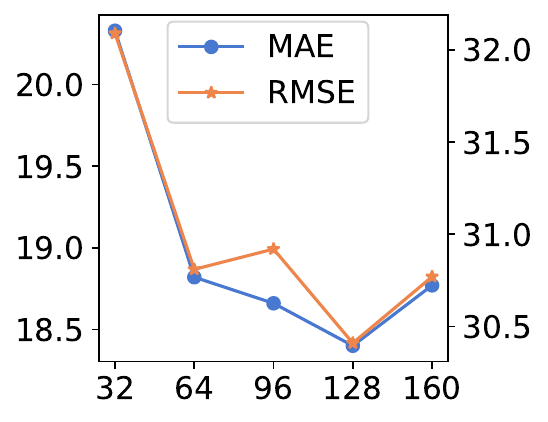}
        \vspace{-6mm}
        \caption{\#Features on PeMS04}
        \label{fig:recall_l}
    \end{subfigure}
    \begin{subfigure}{0.205\textwidth}
        \includegraphics[width=\linewidth]{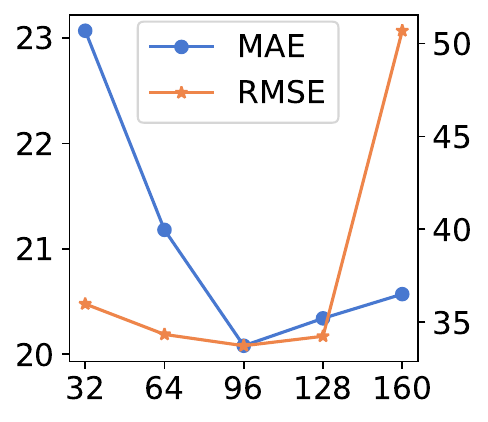}
        \vspace{-6mm}
        \caption{\#Features on PeMS07}
        \label{fig:ndcg_l}
    \end{subfigure}
    \vspace{-1mm}
    \caption{Parameter sensitivity study on PeMS04 and PeMS07 datasets.}
    \label{fig:abla}
    \vspace{-6mm}
\end{figure}

%% file: 6conclusion.tex
\section{Conclusion}
% In this paper, we introduce a novel spatiotemporal-aware trend-seasonality decomposition network, \model, which addresses the complexities of traffic flow prediction by effectively capturing both spatial and temporal dynamics. The model constructs a dynamic graph structure and employs novel spatio-temporal embeddings to accurately represent traffic patterns. These representations are refined through a trend-seasonality decomposition module and processed by an encoder-decoder network, resulting in highly accurate predictions. Extensive experiments on real-world datasets demonstrate STDN's superior performance and efficiency. Additionally, the JiNan traffic dataset, featuring unique inner-city dynamics, has been introduced to enhance traffic prediction evaluation.

% In this paper, we introduce a novel spatiotemporal-aware trend-seasonality decomposition network (STDN), a model that effectively addresses the challenges of traffic flow prediction by capturing both spatial and temporal dynamics. The model constructs a dynamic graph structure and employs spatio-temporal embeddings to divide trend-cyclical part and seasonal part by a trend-seasonality decomposition module and then processed by an encoder-decoder network. Through extensive experiments, STDN demonstrated superior performance and computational efficiency. Additionally, we introduce the JiNan traffic dataset, featuring unique inner-city dynamics.

In this paper, we introduce a novel spatiotemporal-aware trend-seasonality decomposition network (STDN), which marks a pioneering approach in employing spatio-temporal embeddings to learn disentangled representations of traffic flow. The empirical evaluations conducted across three real-world datasets demonstrate the superior performance of \model over existing models. The release of the new inner-city dataset JiNan can also enrich the scenario comprehensiveness in traffic forecasting evaluations.

% we introduce a novel spatiotemporal-aware trend-seasonality decomposition network (STDN), a model that effectively addresses the challenges of traffic flow prediction by capturing both spatial and temporal dynamics. Through extensive experiments, \model demonstrates superior performance and computational efficiency. Additionally, we introduce the JiNan dataset to enrich the scenario comprehensiveness in traffic forecasting.

%% file: 8Acknowledgment.tex
\section*{Acknowledgments}
This work is partially supported by the Fundamental Research Funds for the Central Universities under Grant No 202442005, the National Natural Science Foundation of China under Grant Nos. 62176243 62372421, and 62402463, and the National Key R\&D Program of China under Grant No 2022ZD0117201. 